# Feature Selection in Detection of Adverse Drug Reactions from the Health Improvement Network (THIN) Database

**Yihui Liu**
School of Information Science, Qilu University of Technology, China
Email: yxl@qlu.edu.cn

**Uwe Aickelin**
Department of Computer Science, University of Nottingham, UK
Email: uwe.aickelin@nottingham.ac.uk

*Abstract*—Adverse drug reaction (ADR) is widely concerned for public health issue. ADRs are one of most common causes to withdraw some drugs from market. Prescription event monitoring (PEM) is an important approach to detect the adverse drug reactions. The main problem to deal with this method is how to automatically extract the medical events or side effects from high-throughput medical events, which are collected from day to day clinical practice. In this study we propose a novel concept of feature matrix to detect the ADRs. Feature matrix, which is extracted from big medical data from The Health Improvement Network (THIN) database, is created to characterize the medical events for the patients who take drugs. Feature matrix builds the foundation for the irregular and big medical data. Then feature selection methods are performed on feature matrix to detect the significant features. Finally the ADRs can be located based on the significant features. The experiments are carried out on three drugs: Atorvastatin, Alendronate, and Metoclopramide. Major side effects for each drug are detected and better performance is achieved compared to other computerized methods. The detected ADRs are based on computerized methods, further investigation is needed.

## 1. Introduction

Adverse drug reaction (ADR) is widely concerned for public health issue. ADRs are one of most common causes to withdraw some drugs from market [1]. Now two major methods for detecting ADRs are spontaneous reporting system (SRS) [2-5], and prescription event monitoring (PEM) [6-10]. The World Health Organization (WHO) defines a signal in pharmacovigilance as "any reported information on a possible causal relationship between an adverse event and a drug, the relationship being unknown or incompletely documented previously" [11]. For spontaneous reporting system, many machine learning methods are used to detect ADRs, such as Bayesian confidence propagation neural network (BCPNN) [12，13], decision support method [14], genetic algorithm [15], knowledge based approach [16，17], etc. One limitation is the reporting mechanism to submit ADR reports [14], which has serious underreporting and is not able to accurately quantify the corresponding risk. Another limitation is hard to detect ADRs with small number of occurrences of each drug-event association in the database. Prescription event monitoring has been developed by the Drug Safety Research Unit (DSRU) and is the first large-scale systematic post-marketing surveillance method to use event monitoring in the UK [8]. The health data are widely and routinely collected. It is a key step for PEM methods to automatically detect the ADRs from thousands of medical events. In paper [18，19], MUTARA and HUNT, which are based on a domain-driven knowledge representation Unexpected Temporal Association Rule, are proposed to signal unexpected and infrequent patterns characteristic of ADRs, using Queensland Hospital morbidity data, more commonly referred to as the Queensland Linked Data Set (QLDS) [20]. But their methods achieve low accuracies for detecting ADRs. In the paper [21], four existing ADR detecting algorithms were compared by applying them to The Health Improvement Network (THIN) database for a range of drugs. The results show that the existing algorithms are not capable of detecting rare ADRs. Until now no successful algorithms are proposed to detect the ADRs automatically for PEM methods.

In this paper we propose a novel approach using feature matrix to detect ADRs from The Health Improvement Network (THIN) database. First feature matrix which represents the medical events for the patients before and after taking drugs, is created by linking patients' prescriptions and corresponding medical events together. Then significant features are selected based on feature selection methods, comparing the feature matrix before patients take drugs with one after patients take drugs. Finally the significant ADRs can be detected from thousands of medical events based on corresponding features. Experiments are carried on three drugs of Atorvastatin, Alendronate, and Metoclopramide. Good performance is achieved.

The rest of paper is organized as follows. In section 2, we introduce feature matrix and how to extract it from THIN database. Two feature selection methods are described in section 3. The experiments are conducted in section 4 based on three drugs, followed by discussion in section 5. Conclusions are made in section 6.

## 2. Feature matrix

A novel concept of feature matrix is proposed to represent data and detect ADRs, using feature selection methods. Normally patients take drugs for different periods of time, and have different numbers of repeated prescriptions. The longer the period of time, the more the medical events related to the drug. How to represent the high-throughput data related to prescriptions and medical events, is a key step to detect the ADRs. Feature matrix builds the basis of saving data and comparing data. Two kinds of feature matrix are built in this study, based on medical events using Readcodes at level 1-5 and Readcodes at level 1-3, respectively. The medical events using Readcodes at level 1-3 give the general term, whereas those using Readcodes at level 1-5 give more specific descriptions.

### 2.1 The THIN database

The Health Improvement Network (THIN) is a collaboration product between two companies of EPIC and InPS. EPIC is a research organisation, which provides the electronic database of patient care records from UK and other countries. InPS continue to develop and supply the widely-used Vision general practice computer system. The anonymised patient data are collected from the practice's Vision clinical system on a regular basis without interruption to the running of the GP's system and sent to EPIC who supplies the THIN data to researchers for studies. Research studies for publication using THIN Data are approved by a nationally accredited ethics committee which has also approved the data collection scheme.

There are 'Therapy' and 'Medical' databases in THIN data. The "Therapy" database contains the details of prescriptions issued to patients. Information of patients and the prescription date for the drug can be obtained. The "Medical" database contains a record of symptoms, diagnoses, and interventions recorded by the GP and/or primary care team. Each event for patients forms a record. By linking patient identifier, their prescriptions, and their corresponding medical events or symptoms together, feature matrix to characterize the symptoms during the period before or after patients take drugs is built.

### 2.2 Readcodes and feature matrix

Medical events or symptoms are represented by medical codes, named Readcodes. There are 103387 types of medical events in "Readcodes" database. The Read Codes used in general practice (GP), were invented and developed by Dr James Read in 1982. The NHS (National Health Service) has expanded the codes to cover all areas of clinical practice. The code is hierarchical from left to right or from level 1 to level 5. It means that it gives more detailed information from level 1 to level 5. Table 1 shows the medical symptoms based on Readcodes at level 3 and at level 5. 'Other soft tissue disorders' is general description using Readcodes at level 3. 'Foot pain', 'Heel pain', etc., give more details using Readcodes at level 5. 'Other extrapyramidal disease and abnormal movement disorders' is general term; 'Restless legs syndrome', 'Essential and other specified forms of tremor', etc., are detailed descriptions using Readcodes at level 5.

In this research, two kinds of feature matrix are built to characterize the symptoms of patients who take drugs. One is based on Readcodes at level 1-5, which cover all the symptoms and detailed information which occur when patients take drugs. Another one is based on Readcodes at level 1-3, which is created by combining the detailed symptoms using Readcodes at level 4-5 into the general term using Readcodes at level 3. For the drug of Alendronate, 'Temporomandibular joint disorders' of 'J046.00' at level 4 is a typical ADR, but 'Dentofacial anomalies' of 'J04..00' at level 3 is more general term for this particular ADR. If patients have the symptom of 'Temporomandibular joint disorders' ('J046.00') in feature matrix based Readcodes at 1-5, after we merge the detailed descriptions into the general term, a new feature matrix based on Readcodes at level 1-3 is rebuilt instead of using 'Dentofacial anomalies' ('J04..00').

### 2.3 The extraction of feature matrix

To detect the ADRs of drugs, first feature matrix is extracted from THIN database, which describes the medical events that patients occur before or after taking drugs. Then feature selection methods of Student's t-test and Wilcoxon rank sum test are performed to select the significant features from feature matrix containing thousands of medical events. Figure 1 shows the process to detect the ADRs using feature matrix. Feature matrix $A$ (Figure 2) describes the medical events for each patient during 60 days before they take drugs. Feature matrix $B$ is one after patients take drugs. For repeated prescriptions, all the medical events related to each prescription of patients are input to build the feature matrix. When the interval between two adjacent prescriptions is less than 60 days, we put all the medical events between two adjacent into feature matrix after patients take drugs. When the interval between two prescriptions is more than 60 days, the medical events within 60 days after the first prescription are used build the feature matrix after patients take drugs, and the medical events beyond 60 days are put into the feature matrix before patients take drugs. Figure 3 shows how to extract feature matrix. Figure 4 shows the flowchart of building feature matrix.

In order to reduce the computation time and save the computation space, we set 100 patients as a group. Matrix $X$ and $Y$ are feature matrix after patients are divided into groups.

Feature matrix $A$ and $B$ are defined as follows:

$A = (a_{ij})_{m \times n}, i = 0,1,...,m, j = 1,...,n$ (1)

$a_{ij} \in \{0,1\}$

$B = (b_{ij})_{m \times n}, i = 0,1,...,m, j = 1,...,n$ (2)

$$b_{ij} \in \{0,1\}$$

where variable $i$ and $j$ represent patients and medical events. Variable $m$ and $n$ represent the number of patients and medical events respectively. $a_{ij} = 1$ is for the state where $ith$ patient has $jth$ medical event; otherwise $a_{ij} = 0$. Feature matrix $A$ and $B$ describe the medical events during 60 days before or after patients take drugs.

Feature matrix $X$ and $Y$ are defined as follows:

$$X = (x_{kj})_{g \times n}, \quad k = 0,1,...,g; j = 1,...,n \quad (3)$$

$$x_{kj} = \sum_{i=i_k}^{i_k+d-1} a_{ij},\ i_k = k \times d,\ k = 0,...,g$$

$$Y = (y_{kj})_{g \times n}, \quad k = 0,1,...,g; j = 1,...,n \quad (4)$$

$$y_{kj} = \sum_{i=i_k}^{i_k+d-1} b_{ij},\ i_k = k \times d,\ k = 0,...,g$$

where variable $g$ represents the number of groups. Variable $x_{kj}$ represents the patient number of $kth$ group having $jth$ medical event. In this study, each group contains 100 patients, and $d$ is set to 100.

## 3. Feature selection

After two feature matrixes are obtained, which represent the medical events before and after patients take drugs, feature selection methods are performed to detect the significant features that reflect the significant changes. Feature selection methods normally have two classes. One class is "wrapper" methods, which select feature subsets based on classification performance, such as genetic algorithm [22, 23]. Another one is filter methods, which select features based on between-class discriminant criterion, give the ranked features, and are not involved in classification performance in the process of feature selection, such as Student's t-test and Wilcoxon rank-sum test.

Obviously, in this study, it is not the classification problem. The main idea is to find which medical event changes significantly and give their ranks. We employ feature selection methods of Student's t-test and Wilcoxon rank-sum test, which are based on the data assumption having normal distribution and no assumptions for data distributions, respectively. Based on doctors' suggestion, the ratio of the patient number after taking the drug to one before taking the drug for having one particular symptom is also used to represent the changes of symptoms.

### 3.1 Student's t-test

Student's t-test [24] feature selection method is used to detect the significant ADRs from thousands of medical events. Student's t-test is a kind of statistical hypothesis test based on a normal distribution, and is used to measure the difference between two kinds of samples. Student's t-test calculates a score $t_i$ to measure the difference between feature matrix $X$ and $Y$, which represents the importance of $jth$ medical event or ADR among all groups of patients.

Student's t-test is defined as follows:

$$t_j = \frac{\bar{x}_j - \bar{y}_j}{\sqrt{\frac{S_{xj}^2}{M_x} + \frac{S_{yj}^2}{M_y}}} \quad j = 1,2,...,n \quad (5)$$

where $\bar{x}_j$ and $\bar{y}_j$ represent the means of two groups of samples, $S_{xj}$ and $S_{yj}$ represent the standard deviations of two groups of samples. $M_x$ and $M_y$ represent the number of two groups of samples.

### 3.2 Wilcoxon rank-sum test

Student's t-test assumes that the data follow a normal distribution. If no assumption can be made about the shape of the population distribution, non-parametric statistical methods can be used. The Wilcoxon rank-sum test is a non-parametric analogue to Student's t-test for two independent samples. It is used for the determination of equality of the means of two non-normal samples. The test is based on the rank of the individual data within feature matrix. As it operates on rank-transformed data, it also is a robust choice for detecting the significant medical symptoms. In this test, the ranking of data is performed first by giving the highest rank (equal to the number of features) to the feature with the highest value, and the lowest rank to the feature with lowest value. The ranks for remaining features are assigned using the rank-sum test as explained in details in [25].

### 3.3 Other parameters

The variable of ratio $R_1$ is defined to evaluate significant changes of the medical events, using ratio of the patient number after taking the drug to one before taking the drug. The variable $R_2$ represents the ratio of patient number after taking the drug to the number of whole population having one particular medical symptom.

The ratio variables $R_1$ and $R_2$ are defined as follows:

$$R_1 = \begin{cases} N_A / N_B & \text{if } N_B \neq 0; \\ N_A & \text{if } N_B = 0; \end{cases} \quad (6)$$

$$R_2 = N_A / N$$

where $N_B$ and $N_A$ represent the numbers of patients before or after they take drugs for having one particular medical event respectively. The variable $N$ represents the number of whole population who take drugs.

We compare our proposed method with published results in [19]. Same evaluation standard is used in our research. Accuracy is defined as follows:

$$ACC = N_{adr} / N_{all}$$

where *ACC* represents the accuracy to predict the ADRs. $N_{adr}$ represents the number of true positive (ADRs), and $N_{all}$ represents the number of whole number for prediction. In this research we only evaluate top 20 results of prediction, and $N_{all}$ is set to 20.

## 4. Experiments and results

Three drugs of Atorvastatin, Alendronate, and Metoclopramide are used to test our proposed method, using 20 GPs data in THIN database. The drug of Atorvastatin is one of 'statin' class and has the particular 'muscle pain' side effects. For the drug of Alendronate, the 'Temporomandibular joint disorders' is a typical ADR. The drug of Metoclopramide has the typical ADR of 'extrapyramidal effects' or 'abnormal movement disorders'. So in this study, three drugs that have different typical ADRs, are used to test our proposed method. Jin et al. also use the drugs of Atorvastatin and Alendronate to test their proposed methods of MUTARA and HUNT [18,19].

Student's t-test and Wilcoxon rank-sum test are performed to select the significant features from feature matrix, which represent the medical events having the significant changes after patients take the drugs. In experiments, we use two kinds of feature matrix. One feature matrix is based on all medical events using Readcodes at level 1-5 in order to observe the detailed symptoms. Another one is based on the medical events using Readcodes at level 1-3, by combining the detailed information of Readcodes at level 4 and 5 into general terms using Readcodes at level 3.

Some common abbreviations used in the 'Readcodes' dictionary are shown as below:
C/O    Complains of
H/O    History of
NOS    Not otherwise specified
F/H    Family history
O/E    on examination
[M]    Morphology of neoplasms
[D]    Working diagnosis

**4.1 Atorvastatin**
Atorvastatin [26], under the trade name Lipitor, is one of the drugs known as statins. It is used for lowering blood cholesterol. There are seven currently prescribed forms of statin drugs. They are Rosuvastatin, Atorvastatin, Simvastatin, Pravastatin, Fluvastatin, Lovastatin, Pitavastatin. Muscle pain and musculoskeletal events are two of the main side effects of statin drugs [27].

Drugs.com provides access to healthcare information, sourced solely from the most trusted, well-respected and independant agents such as the Food and Drug Administration (FDA), American Society of Health-System Pharmacists, Wolters Kluwer Health, Thomson Micromedex, Cerner Multum and Stedman's. The side effects for Atorvastatin [28] by Drugs.com are severe allergic reactions (rash; hives; itching; difficulty breathing or swallowing; tightness in the chest; swelling of the mouth, face, lips, throat, or tongue); burning, numbness, or tingling; change in the amount of urine produced; confusion; memory problems; mental or mood problems (eg, depression); muscle pain, tenderness, or weakness (with or without fever or fatigue); painful, difficult, or frequent urination; persistent pain, soreness, redness, or swelling of a tendon or joint; red, swollen, blistered, or peeling skin; severe stomach or back pain (with or without nausea or vomiting); symptoms of liver problems (eg, dark urine; pale stools; severe or persistent nausea, loss of appetite, or stomach pain; unusual tiredness; yellowing of the skin or eyes).

We only use the data for patients, who registered on GP for more than one year, in order to obtain the reliable data. 6803 patients take Atorvastatin from 20GP data in THIN database. Based on Readcodes at level 1-5, totally 10528 medical events are obtained before or after 6803 patients take the drug. So 6803x10528 feature matrix is obtained. Based on Readcodes at level 1-3, we combine the medical events which have the same first three codes into one medical event, just using first three codes. 10528 medical events based on Readcodes at level 1-5 are combined into 2350 medical events based on Readcodes at level 1-3, and a 6803x2350 feature matrix is created. After grouping patients, 68x10528 and 68x2350 feature matrix are formed to select the significant features, which reflect the significant change of medical events after patients take drugs.

Table 2 and Table 3 show the top 20 ADRs for Atorvastatin based on Student's t-test and Wilcoxon rank-sum test, using medical events based on Readcodes at level 1-5 and level 1-3. From Table 2 and 3, it is clear that the detected ADRs are consistent with the published side effects for Atorvastatin. The difference for the detected ADRs between Student's t-test and Wilcoxon rank-sum test is that the rank of the ADRs is different. For example the medical event of 'Other soft tissue disorder' is rank 3 based on Student's t-test, and rank 12 based on Wilcoxon rank-sum test. 'Other soft tissue disorder' is major side effect for statin. From this view, the detected ADRs based on Student's t-test are better than those based on Wilcoxon rank-sum test.

Our detected ADRs are classified into several categories, such as muscle pain, muscle weakness and neuropathy, gastrointestinal events, musculoskeletal events [27], skin and subcutaneous tissue events [28], bronchitis and bronchiolitis [29] diabetes mellitus events [30，31]. Good performance of 100% accuracy is achieved for the top 20 ADRs based on Student's t-test and Wilcoxon rank-sum test.

MUTARA and HUNT [18，19] based on Unexpected Temporal Association Rule are proposed to signal unexpected and infrequent patterns characteristic of ADRs, using The Queensland Linked Data Set (QLDS). They indicated that "HUNT can reliably shortlist statistically significantly more ADRs than MUTARA". For the drug of Atorvastatin, HUNT detects 4 ADRs of

'urinary tract infection', 'stomach ulcer', 'diarrhoea', and 'bronchitis' from top 20 results based on 13712 patient records, and only obtains 30% average accuracy [19]. The major side effects of Atorvastatin, such as muscle pain and musculoskeletal events, are not detected. Table 4 shows the performance of HUNT and MUTARA. Our proposed method detects 'muscle and musculoskeletal' events not only based on Readcodes at level 1-5, but also based on Readcodes at level 1-3. 'Peripheral enthesopathies and allied syndromes' and 'Other soft tissue disorders' are detected at rank 2 and 3 based on Readcodes at level 1-3, using feature selection method of Student's t-test.

### 4.2 Alendronate

Alendronate sodium, sold as Fosamax by Merck, is a bisphosphonate drug used for osteoporosis and several other bone disease [32]. The side effects for Alendronate [33] are severe allergic reactions (rash; hives; itching; difficulty breathing; tightness in the chest; swelling of the mouth, face, lips, throat, or tongue); black, tarry, or bloody stools; chest pain; coughing or vomiting blood; difficult or painful swallowing; mouth sores; new, worsening, or persistent heartburn; red, swollen, blistered, or peeling skin; severe bone, muscle, or joint pain (especially in the hip, groin, or thigh); severe or persistent sore throat or stomach pain; swelling of the hands, legs, or joints; swelling or pain in the jaw; symptoms of low blood calcium (eg, spasms, twitches, or cramps in your muscles; numbness or tingling in your fingers, toes, or around your mouth).

In [32], side effects include ulceration of the esophagus, gastric and duodenal ulceration, skin rash, rarely manifesting as Stevens–Johnson syndrome and toxic epidermal necrolysis, eye problems (uveitis, scleritis), generalized muscle, joint, and bone pain, osteonecrosis of the jaw or deterioration of the Temporomandibular Joint (TMJ), auditory hallucinations, visual disturbances. Studies suggest that users of alendronate have an increase in the numbers of osteoclasts and develop giant, more multinucleated osteoclasts. Fosamax has been linked to a rare type of leg fracture that cuts straight across the upper thigh bone after little or no trauma.

3346 patients from 20GP data in THIN database are taking Alendronate, and 7260 medical events are obtained based on Readcodes at level 1-5. After grouping them, 33x7260 feature matrix is obtained. For Readcodes at level 1-3, 33x1964 feature matrix is obtained. Table 5 and 6 show the top 20 ADRs based on Student's t-test and Wilcoxon rank-sum test. Our results show that most detected ADRs are consistent with the published results [32]. 'Chronic kidney disease stage 3', 'Type 2 diabetes mellitus', and 'Essential hypertension' are not listed in [32]. 85% accuracy is achieved. But in [34], 'acute renal failure' is reported after the patient takes Alendronate. 'Based on the study created by eHealthMe from U.S. Food and Drug Administration (FDA) and user community [35], 0.60% patients have 'Type 2 diabetes mellitus'. In [36], 0.62% patients have 'Essential hypertension'. For HUNT, only 'headache', 'swelling', 'wheeze', 'breathing difficult', 'heart burn', and 'constipation' are detected in top 20 ADRs based on 7569 patients. Totally 30% average accuracy is obtained. The major side effect of 'muscle or joint pain' is not detected.

'Peripheral enthesopathies and allied syndromes' and 'Other soft tissue disorders' with rank 14 and 15 are detected to describe the ADRs related to 'muscle or joint pain', which are listed in Table 5 using Readcodes at level 1-3 based on Student's t-test. But they are not list in Table 6 based on Wilcoxon rank-sum test in top 20 ADRs. From this view, Student's t-test method is better than Wilcoxon rank-sum test.

After we obtain the results using Student's t-test with p<0.05, which reflect the significant changes after patients take drug, then we also sort the order of results according to descending order of R1 value. Table 7 shows top 20 ADRs based on the descending order of R1 value. From Table 7, we notice that the typical side effect 'Temporomandibular joint disorders' for Alendronate' is detected at rank 11 using Readcodes at level 1-5.

### 4.3 Metoclopramide

Metoclopramide is used to treat nausea and vomiting associated with conditions such as uremia, radiation sickness, malignancy, labor, infection, migraine headaches, and emetogenic drugs [37]. The side effects [38] for Metoclopramide are severe allergic reactions (rash; hives; itching; difficulty breathing; tightness in the chest; swelling of the mouth, face, lips, or tongue; unusual hoarseness); abnormal thinking; confusion; dark urine; decreased balance or coordination; decreased sexual ability; fast, slow, or irregular heartbeat; fever; hallucinations; loss of bladder control; mental or mood changes (eg, depression, anxiety, agitation, jitteriness); seizures; severe or persistent dizziness, headache, or trouble sleeping; severe or persistent restlessness, including inability to sit still; shortness of breath; stiff or rigid muscles; sudden increased sweating; sudden, unusual weight gain; suicidal thoughts or actions; swelling of the arms, legs, or feet; uncontrolled muscle spasms or movements (eg, of the arms, legs, tongue, jaw, cheeks; twitching; tremors); vision changes; yellowing of the skin or eyes. In [37] common adverse drug reactions associated with metoclopramide therapy include restlessness, drowsiness, dizziness, fatigue, and focal dystonia. Infrequent ADRs include hypertension, hypotension, hyperprolactinaemia leading to galactorrhea, constipation, depression, headache, and extrapyramidal effects such as oculogyric crisis.

8320 patients from 20 GPs take Metoclopramide. 8094 medical events are obtained using Readcodes at level 1-5, and 8320x8094 feature matrix is created. After 8094 medical events are combined into 2117 medical symptoms based on Readcodes at level 1-3, 8320x2117 feature matrix is created. After patients are grouped, 83x8094 feature matrix is formed using Readcodes at level 1-5, and 83x2117 feature matrix is obtained using Readcodes at level 1-3. Table 8 and 9 show the top 20 detected results in ascending order of p value of Student's t-test and Wilcoxon rank-sum test, using Readcodes at level 1-5 and at level 1-3, respectively.

From Table 8 and 9, it is shown that the typical side effect 'Other extrapyramidal disease and abnormal movement disorders' for Metoclopramide is detected at rank 3 and rank 4 based on Student's t-test and Wilcoxon rank-sum test. The top 20 medical events cover most important side effects for the drug of Metoclopramide. The side effect 'Cholangitis' is detected at rank 14 in Table 8 based on Student's t-test using Readcodes at level 1-5. In [38], a 22-year-old female, treated with metoclopramide for 7 to 8 months for abdominal pain, developed hepatic hemangiomatosis with arteriovenous shunting and cholestasis.

The ADRs having top ranks (rank 1,2,3,5) are 'patient died' based on Student's t-test using Readcodes at level 1-5, and they are not listed in [37, 38]. So our result achieves 80% accuracy. But based on the study created by eHealthMe from FDA and user community, 1.93% patients have 'Death' [39]. We notice that three patients are 'died' before they took the drug. Actually, the real state for these patients is 'expected died'.

## 5. Discussion

### 5.1 Feature matrix

In this research, we propose the novel concept of feature matrix. For the medical data which are collected from day to day clinical practice, they are irregular and magnanimity data. Each patient has different medical information. Even for the same drug, different patients have taken drugs at different time points and different periods, and have different medical symptoms. How to extract the useful information after patients take drugs is first and key step to detect the adverse drug reactions. A novel concept of feature matrix is proposed to store and characterize the common information that only deal with symptoms before and after taking drugs for patients. For each drug, its corresponding feature matrix is built to cover all the symptoms or medical events occurred before or after taking the drug for all patients. Feature matrix transfers the unrelated information of different patients to common saving format, and builds the foundation to detect the adverse drug reactions. For example, some patients take the "statin" drugs for years, and the corresponding prescriptions and medical events are "mass" information. But we only interest the information before and after taking drugs, feature matrix are built to characterize the representative features extracted from THIN database, which is collected day to day clinical practice. All the prescriptions during the period of taking drugs are checked. The medical events during 60 days before or after each prescription are used to build the feature matrix for this drug. This is reason that chronic side effects can be detected. For example, for the drug of Alendronate, one chronic side effect is "Temporomandibular joint disorders", which is detected by our proposed method.

Two kinds of feature matrix are created to represents the medical events or symptoms for patients before and after taking drugs. One type of feature matrix is to use original data based on Readcodes at level 1-5. Another one is based on Readcodes at level 1-3, combining the medical symptoms using Readcodes at level 4-5 into one using Readcodes at level 3. For example, 'Type 2 diabetes mellitus' has Readcodes of 'C10F.00' at level 4, and Readcodes of 'C10..00' at level 3 represents the general term of 'Diabetes mellitus'.

### 5.2 Feature selection and the performance

In this research, we compare our proposed method with published results in [19]. In paper [19], two drugs of Atovastatin and Alendronate are used to test their proposed methods, and average 30% accuracy is obtained in [19]. Their proposed methods do not detect two major side effects of muscle pain and musculoskeletal events.

Same evaluation standard is used in our research. Our proposed methods based on extracted feature matrix and feature selection, achieve good results. For drug Atovastatin and Alendronate, major ADRs are detected, and 100% and 85% accuracy are obtained. For Atorvastatin, the major side effect of 'muscle pain' is detected at rank 3, which is 'Other soft tissue disorders' using Readcodes at level 1-3. For the drug of Alendronate, the typical side effect of 'Temporomandibular joint disorders' is detected at rank 1 based on the descending order of the ratio of the patient number after taking the drug to one before taking the drug, which is 'Dentofacial anomalies' using Readcodes at level 1-3. For the drug of Metoclopramide, the typical ADR of 'abnormal movement disorders' is detected at rank 3, which is 'Other extrapyramidal disease and abnormal movement disorders' using Readcodes at level 1-3.

Feature selection is widely used to reduce data dimensionality and detect the significant features [40,41]. In this research feature selection methods of Student's t-test and Wilcoxon rank-sum test are employed to detect the significant changes of feature matrix. Student's t-test assumes that the data follow a normal distribution. Wilcoxon rank-sum test has no assumption about the shape of data distribution. Experimental results show that Student's t-test achieves better performance than Wilcoxon rank-sum test. The variable of ratio of the patient number after taking the drug to one before taking the drug is complementary part for Student's t-test method. The ratio variable is useful to detect the some typical ADRs.

Because of the creation of feature matrix, its unique saving format makes it possible to combine medical symptoms using Readcodes at level 4-5 into one using Readcodes at level 3. Feature matrix using Readcodes at level 1-5 gives the detail information, and feature matrix using Readcode at level 1-3 can extracts the general symptoms, which may be not noticed using feature matrix using Readcodes at level 1-5, for example the ADR of "extrapyramidal disease and abnormal movement disorder" for drug Metoclopramide.

### 5.3 Rare ADRs, chronic ADRs, drug interaction, and advantages

Our proposed method has successfully shortlisted the rare ADRs of "Temporomandibular joint disorders" for drug Alendronate, "extrapyramidal disease and abnormal movement disorder" for drug Metoclopramide, and the research work in [21] indicated that the existing algorithms are not capable of detecting rare ADRs.

The ADRs are rare, and it means the frequency to appear is low. If the methods detect the ADRs only based on the frequency of ADR appearing, there is never sufficient data to support the hypothesis of a causal relationship between the drug and the reaction.

The rare ADRs of "Temporomandibular joint disorders" for drug Alendronate and "extrapyramidal disease and abnormal movement disorder" for drug Metoclopramide are from 3346 and 8320 patients. Enough data is used to support the detection of ADRs. Our proposed method is not only based on the frequency of ADR appearing, and but also is based on the significant change of medical symptoms. The rare ADRs may not appear frequently, but make the significant change. So our proposed method can detect rare ADRs.

When feature matrix is built, all the prescriptions of each patient for this drug are checked to extract the information, no matter how long the patient takes this drug. Because of the characteristics of feature matrix, chronic side effects can be detected. Exiting algorithms [18，19] only check the prescriptions during some period of time, and lose part of the information of side effects.

Because of the characteristics of feature matrix, it is easy to trace the information of patients having the particular ADRs, and extract the information of other drugs taken by these patients. This gives the foundation to do further research of drug interactions.

Two advantages are achieved in this research compared with exiting algorithms. One is to propose feature matrix, which transfers the irregular information of each patient during whole period of taking drugs into standard saving format. This makes it possible to detect chronic ADRs. Another one is based on the changing degree of symptoms to detect the ADRs. This makes it possible to detect rare ADRs.

## 6. Conclusions

In this study we propose a novel method to successfully detect the ADRs by introducing feature matrix and feature selection to detect the significant changes after patients take drugs. A feature matrix, which characterizes the medical events before patients take drugs or after patients take drugs, is created from THIN database. Feature matrix transfers the irregular and high-throughput medical data collected from daily basis into feature matrix of standard saving format, and is a foundation to perform feature selection methods. Feature selection methods based on Student's t-test and Wilcoxon rank-sum test are used to detect the significant features from high dimensional feature matrix. The significant ADRs, which are corresponding to significant features, are detected. Experiments are performed on three drugs: Atorvastatin, Alendronate, and Metoclopramide. Compared to other computerized method, our proposed method achieves better performance. The detected ADRs are based on computerized methods, further investigation is needed.

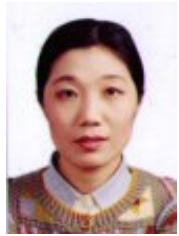

**Authors' Profiles**

Yihui Liu is Professor at Qilu University of Technology, Shandong, China. She received her PhD degree from University of Nottingham, UK, 2004. She was doing postdoctoral research and international cooperation project from 2004 to 2005, and from 2008 to 2012, at University of Nottingham and Aberystwyth University, UK. Her research interests focus on biomedical data analysis, high dimensional data analysis, and wavelet feature extraction, etc.

Uwe Aickelin is EPSRC Advanced Research Fellow and Professor of Computer Science at the University of Nottingham, where he leads one of its four research groups: Intelligent Modelling & Analysis (IMA). His long-term research vision is to create an integrated framework of problem understanding, modelling and analysis techniques, based on an inter-disciplinary perspective of their closely-coupled nature. A summary of his current research interests is Modelling, Artificial Intelligence and Complexity Science for Data Analysis.

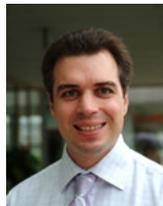

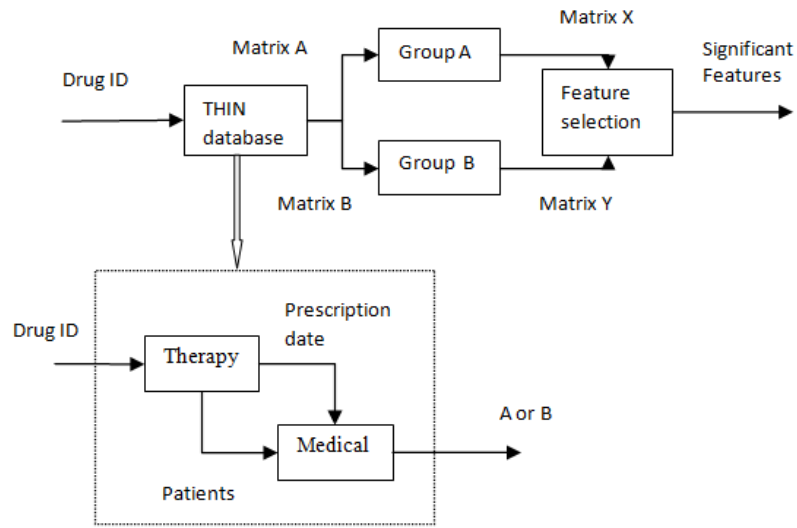

Fig. 1. The process to detect ADRs. Matrix $A$ and $B$ are feature matrix before or after patients take drugs. The time period of observation is set to 60 days. Matrix $X$ and $Y$ are feature matrix after patients are divided into groups. We set 100 patients as one group.

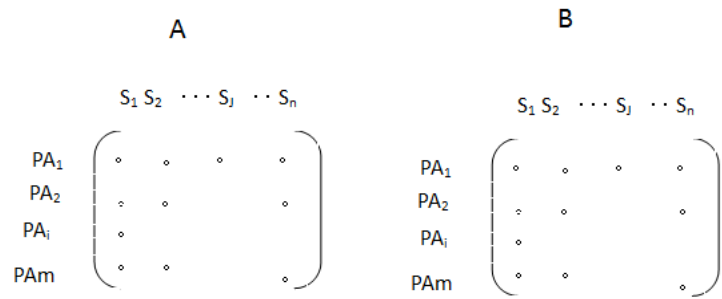

Fig. 2. Feature matrix. The row of feature matrix represents the patients, and the column of feature matrix represents the medical events. The element of events is 1 or 0, which represents that patients have or do not have the corresponding medical symptom or event.

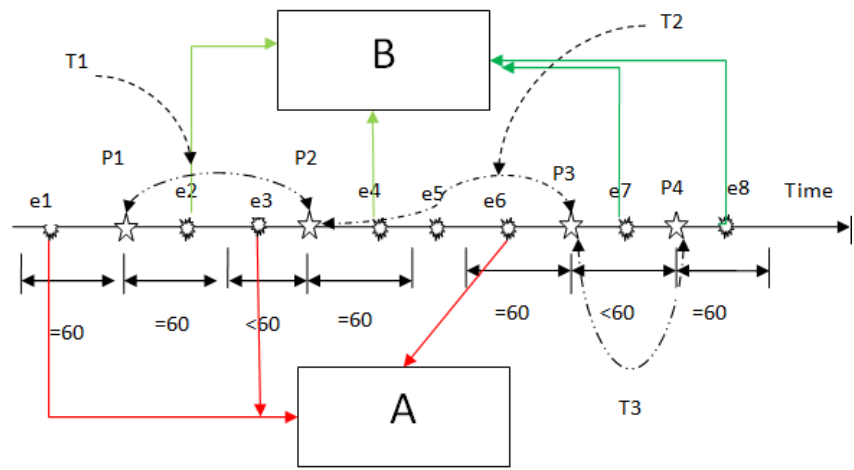

Fig.3. The building process of feature matrix. Feature matrix A reflects the medical events for each patient during 60 days before they take drugs. Feature matrix B is one after patients take drugs. Time axis covers time points of prescriptions and medical events during the whole time for the patients taking drug. Variable p1,p2,p3 and p4 (☆) represent the time points of prescriptions for the patient taking the drug. Variable e1,e2,..., and e8 (○) represent the time points of medical symptoms for the patient. The interval T1 between p1 and p2 is 60<T1<120 days. The interval T2 between p2 and p3 is T2>120 days. The interval T3 between p3 and p4 is <=60 days.

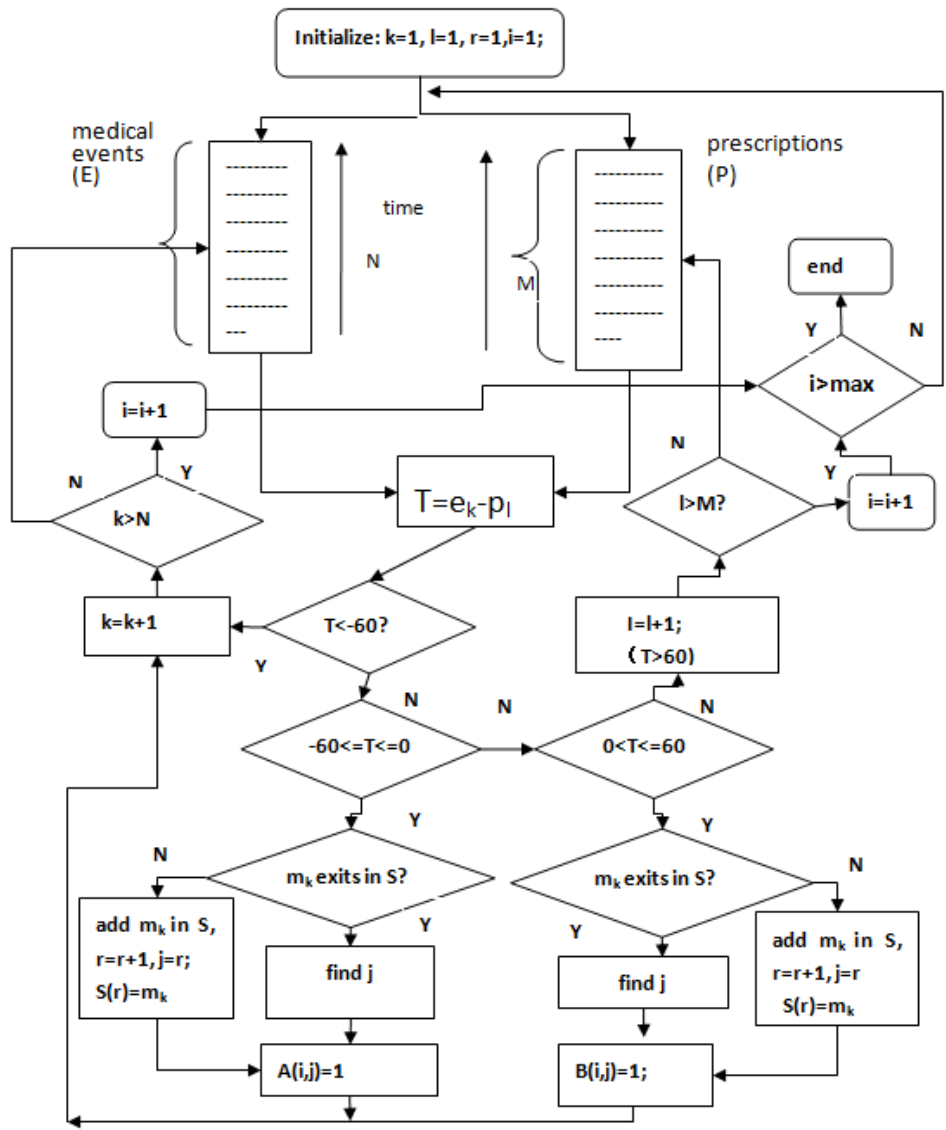

Fig. 4. The flowchart of building feature matrix. Feature matrix A reflects the medical events for each patient during 60 days before they take drugs. Feature matrix B is one after patients take drugs. Two datasets of medical events E and prescriptions P include all the medical events and prescriptions of patients taking the drug. The datasets of E and P are extracted from THIN database, and all the data are sorted in ascending order along the time. Variable $p_l$ and $e_k$ represent the time points for the prescription (l) and the medical event (k) respectively. Variable k (k=1,...,N) and l (l=1,...,M) represent the number or position of the medical event in dataset E and the prescriptions in dataset P respectively. Variable S represents the dataset including all the types of medical events in A and B. Variable r represents the type number of medical events in S. Variable i and j represent ith patient and jth medical event respectively. Variable $m_k$ represents the kth medical event in dataset E.

Table 1. Medical events based on Readcodes at level 1-3 and level 1-5.

| | Level | Readcodes | Medical events |
|---|---|---|---|
| Other soft tissue disorders | Level 3 | N24..00 | Other soft tissue disorders |
| | Level 5 | N245.16 | Leg pain |
| | | N245111 | Toe pain |
| | | N241012 | Muscle pain |
| | | N242000 | Neuralgia unspecified |
| | | N241000 | Myalgia unspecified |
| Abnormal movement | Level 3 | F13..00 | Other extrapyramidal disease and abnormal movement disorders |
| | Level 5 | F13z200 | Restless legs syndrome |
| | | F131000 | Benign essential tremor |
| | | F131z00 | Essential and other specified forms of tremor |
| | | F132z12 | Myoclonic seizure |
| | | F132300 | Myoclonic jerks |

Table 2. The top 20 ADRs for Atorvastatin based on Student's t-test.

| | Rank | Readcodes | Medical events | NB | NA | R1(%) | R2 (%) |
|---|---|---|---|---|---|---|---|
| Level 1-5 | 1 | 1A55.00 | Dysuria | 26 | 181 | 6.96 | 2.66 |
| | 2 | 1Z12.00 | Chronic kidney disease stage 3 | 63 | 477 | 7.57 | 7.01 |
| | 3 | N131.00 | Cervicalgia - pain in neck | 73 | 337 | 4.62 | 4.95 |
| | 4 | N143.00 | Sciatica | 48 | 205 | 4.27 | 3.01 |
| | 5 | F4C0.00 | Acute conjunctivitis | 60 | 274 | 4.57 | 4.03 |
| | 6 | N245.17 | Shoulder pain | 99 | 376 | 3.80 | 5.53 |
| | 7 | M03z000 | Cellulitis NOS | 57 | 264 | 4.63 | 3.88 |
| | 8 | 1C14.00 | "Blocked ear" | 13 | 120 | 9.23 | 1.76 |
| | 9 | A53..11 | Shingles | 17 | 121 | 7.12 | 1.78 |
| | 10 | H01..00 | Acute sinusitis | 44 | 194 | 4.41 | 2.85 |
| | 11 | 1M10.00 | Knee pain | 103 | 369 | 3.58 | 5.42 |
| | 12 | 19EA.00 | Change in bowel habit | 17 | 102 | 6.00 | 1.50 |
| | 13 | H06z000 | Chest infection NOS | 136 | 563 | 4.14 | 8.28 |
| | 14 | C10F.00 | Type 2 diabetes mellitus | 154 | 436 | 2.83 | 6.41 |
| | 15 | 1D14.00 | C/O: a rash | 70 | 357 | 5.10 | 5.25 |
| | 16 | M0...00 | Skin and subcutaneous tissue infections | 18 | 115 | 6.39 | 1.69 |
| | 17 | N247100 | Leg cramps | 36 | 132 | 3.67 | 1.94 |
| | 18 | M111.00 | Atopic dermatitis/eczema | 36 | 190 | 5.28 | 2.79 |
| | 19 | 19F..00 | Diarrhoea symptoms | 52 | 244 | 4.69 | 3.59 |
| | 20 | H33..00 | Asthma | 23 | 106 | 4.61 | 1.56 |
| Level 1-3 | 1 | 171..00 | Cough | 309 | 1105 | 3.58 | 16.24 |
| | 2 | N21..00 | Peripheral enthesopathies and allied syndromes | 136 | 568 | 4.18 | 8.35 |
| | 3 | N24..00 | Other soft tissue disorders | 462 | 1371 | 2.97 | 20.15 |
| | 4 | 1Z1..00 | Chronic renal impairment | 79 | 570 | 7.22 | 8.38 |
| | 5 | 1B1..00 | General nervous symptoms | 204 | 667 | 3.27 | 9.80 |
| | 6 | 19F..00 | Diarrhoea symptoms | 97 | 444 | 4.58 | 6.53 |
| | 7 | M22..00 | Other dermatoses | 94 | 353 | 3.76 | 5.19 |
| | 8 | 1C1..00 | Hearing symptoms | 43 | 269 | 6.26 | 3.95 |
| | 9 | 183..00 | Oedema | 97 | 433 | 4.46 | 6.36 |
| | 10 | N13..00 | Other cervical disorders | 81 | 362 | 4.47 | 5.32 |
| | 11 | 1B8..00 | Eye symptoms | 73 | 322 | 4.41 | 4.73 |
| | 12 | F4C..00 | Disorders of conjunctiva | 85 | 381 | 4.48 | 5.60 |
| | 13 | H06..00 | Acute bronchitis and bronchiolitis | 357 | 1161 | 3.25 | 17.07 |
| | 14 | 173..00 | Breathlessness | 198 | 675 | 3.41 | 9.92 |
| | 15 | 1A5..00 | Genitourinary pain | 56 | 276 | 4.93 | 4.06 |
| | 16 | M03..00 | Other cellulitis and abscess | 73 | 357 | 4.89 | 5.25 |
| | 17 | N09..00 | Other and unspecified joint disorders | 196 | 721 | 3.68 | 10.60 |
| | 18 | 1D1..00 | C/O: a general symptom | 159 | 695 | 4.37 | 10.22 |
| | 19 | A53..00 | Herpes zoster | 20 | 158 | 7.90 | 2.32 |
| | 20 | J57..00 | Other disorders of intestine | 48 | 222 | 4.63 | 3.26 |

Variable NB and NA represent the numbers of patients before or after they take drugs for having medical events. Variable R1 represents the ratio of the numbers of patients after taking drugs to the numbers of patients before taking drugs for having one particular ADR. Variable R2 represents the ratio of the numbers of patients after taking drugs to the number of the whole population for having one particular ADR.

Table 3. The top 20 ADRs for Atorvastatin based on Wilcoxon rank-sum test.

| | Rank | Readcodes | Medical events | NB | NA | R1(%) | R2 (%) |
|---|---|---|---|---|---|---|---|
| Level 1-5 | 1 | 1Z12.00 | Chronic kidney disease stage 3 | 63 | 477 | 7.57 | 7.01 |
| | 2 | 1A55.00 | Dysuria | 26 | 181 | 6.96 | 2.66 |
| | 3 | N131.00 | Cervicalgia - pain in neck | 73 | 337 | 4.62 | 4.95 |
| | 4 | M03z000 | Cellulitis NOS | 57 | 264 | 4.63 | 3.88 |
| | 5 | C10F.00 | Type 2 diabetes mellitus | 154 | 436 | 2.83 | 6.41 |
| | 6 | A53..11 | Shingles | 17 | 121 | 7.12 | 1.78 |

|  | 7 | N143.00 | Sciatica | 48 | 205 | 4.27 | 3.01 |
|---|---|---|---|---|---|---|---|
|  | 8 | F4C0.00 | Acute conjunctivitis | 60 | 274 | 4.57 | 4.03 |
|  | 9 | 1C14.00 | "Blocked ear" | 13 | 120 | 9.23 | 1.76 |
|  | 10 | N245.17 | Shoulder pain | 99 | 376 | 3.80 | 5.53 |
|  | 11 | H01..00 | Acute sinusitis | 44 | 194 | 4.41 | 2.85 |
|  | 12 | K190.00 | Urinary tract infection, site not specified | 56 | 271 | 4.84 | 3.98 |
|  | 13 | 19EA.00 | Change in bowel habit | 17 | 102 | 6.00 | 1.50 |
|  | 14 | M0...00 | Skin and subcutaneous tissue infections | 18 | 115 | 6.39 | 1.69 |
|  | 15 | M223.00 | Seborrhoeic keratosis | 15 | 99 | 6.60 | 1.46 |
|  | 16 | H33..00 | Asthma | 23 | 106 | 4.61 | 1.56 |
|  | 17 | N245.13 | Foot pain | 36 | 165 | 4.58 | 2.43 |
|  | 18 | C34..00 | Gout | 66 | 194 | 2.94 | 2.85 |
|  | 19 | G573000 | Atrial fibrillation | 51 | 144 | 2.82 | 2.12 |
|  | 20 | M21z100 | Skin tag | 14 | 86 | 6.14 | 1.26 |
| Level 1-3 | 1 | 1C1..00 | Hearing symptoms | 43 | 269 | 6.26 | 3.95 |
|  | 2 | 1Z1..00 | Chronic renal impairment | 79 | 570 | 7.22 | 8.38 |
|  | 3 | H06..00 | Acute bronchitis and bronchiolitis | 357 | 1161 | 3.25 | 17.07 |
|  | 4 | M03..00 | Other cellulitis and abscess | 73 | 357 | 4.89 | 5.25 |
|  | 5 | N21..00 | Peripheral enthesopathies and allied syndromes | 136 | 568 | 4.18 | 8.35 |
|  | 6 | J57..00 | Other disorders of intestine | 48 | 222 | 4.63 | 3.26 |
|  | 7 | 171..00 | Cough | 309 | 1105 | 3.58 | 16.24 |
|  | 8 | A53..00 | Herpes zoster | 20 | 158 | 7.90 | 2.32 |
|  | 9 | 173..00 | Breathlessness | 198 | 675 | 3.41 | 9.92 |
|  | 10 | M22..00 | Other dermatoses | 94 | 353 | 3.76 | 5.19 |
|  | 11 | K19..00 | Other urethral and urinary tract disorders | 101 | 496 | 4.91 | 7.29 |
|  | 12 | N24..00 | Other soft tissue disorders | 462 | 1371 | 2.97 | 20.15 |
|  | 13 | 19F..00 | Diarrhoea symptoms | 97 | 444 | 4.58 | 6.53 |
|  | 14 | 183..00 | Oedema | 97 | 433 | 4.46 | 6.36 |
|  | 15 | F42..00 | Other retinal disorders | 65 | 320 | 4.92 | 4.70 |
|  | 16 | 1B1..00 | General nervous symptoms | 204 | 667 | 3.27 | 9.80 |
|  | 17 | F4C..00 | Disorders of conjunctiva | 85 | 381 | 4.48 | 5.60 |
|  | 18 | 1A5..00 | Genitourinary pain | 56 | 276 | 4.93 | 4.06 |
|  | 19 | H01..00 | Acute sinusitis | 53 | 236 | 4.45 | 3.47 |
|  | 20 | 1B8..00 | Eye symptoms | 73 | 322 | 4.41 | 4.73 |

Table 4. The accuracy of HUNT and MUTARA [19].

| Experimental settings | | Signalling accuracy | |
|---|---|---|---|
| Drug | Patients | HUNT | MUTARA |
| Atorvastatin | Older female | 0.30 | 0.20 |
|  | Older male | 0.30 | 0.20 |
|  | All patients (13712) | 0.30 | 0.20 |
| Alendronate | Older female | 0.50 | 0.30 |
|  | Older male | 0.40 | 0.20 |
|  | All patients (7569) | 0.30 | 0.20 |

Table 5. The top 20 ADRs for Alendronate based on Student's t-test.

|  | Rank | Readcodes | Medical events | NB | NA | R1(%) | R2 (%) |
|---|---|---|---|---|---|---|---|
| Level 1-5 | 1 | 1Z12.00 | Chronic kidney disease stage 3 | 51 | 189 | 3.71 | 5.65 |
|  | 2 | F4C0.00 | Acute conjunctivitis | 30 | 125 | 4.17 | 3.74 |
|  | 3 | M03z000 | Cellulitis NOS | 37 | 152 | 4.11 | 4.54 |
|  | 4 | 1M10.00 | Knee pain | 43 | 154 | 3.58 | 4.60 |
|  | 5 | C10F.00 | Type 2 diabetes mellitus | 7 | 44 | 6.29 | 1.32 |
|  | 6 | H06z000 | Chest infection NOS | 86 | 297 | 3.45 | 8.88 |
|  | 7 | J16y400 | Dyspepsia | 29 | 104 | 3.59 | 3.11 |
|  | 8 | M0...00 | Skin and subcutaneous tissue infections | 9 | 65 | 7.22 | 1.94 |
|  | 9 | 182..00 | Chest pain | 73 | 203 | 2.78 | 6.07 |
|  | 10 | 173I.00 | MRC Breathlessness Scale: grade 2 | 9 | 43 | 4.78 | 1.29 |
|  | 11 | N05..11 | Osteoarthritis | 39 | 138 | 3.54 | 4.12 |
|  | 12 | A53..11 | Shingles | 15 | 56 | 3.73 | 1.67 |
|  | 13 | N131.00 | Cervicalgia - pain in neck | 48 | 132 | 2.75 | 3.95 |
|  | 14 | G20..00 | Essential hypertension | 30 | 116 | 3.87 | 3.47 |
|  | 15 | 1C14.00 | "Blocked ear" | 5 | 39 | 7.80 | 1.17 |
|  | 16 | 1A55.00 | Dysuria | 21 | 76 | 3.62 | 2.27 |
|  | 17 | F46..00 | Cataract | 14 | 67 | 4.79 | 2.00 |
|  | 18 | K190.00 | Urinary tract infection, site not specified | 62 | 214 | 3.45 | 6.40 |
|  | 19 | N143.00 | Sciatica | 33 | 103 | 3.12 | 3.08 |
|  | 20 | AB2..12 | Thrush | 14 | 59 | 4.21 | 1.76 |
|  | 1 | 1Z1..00 | Chronic renal impairment | 58 | 237 | 4.09 | 7.08 |
|  | 2 | 183..00 | Oedema | 77 | 252 | 3.27 | 7.53 |
|  | 3 | C10..00 | Diabetes mellitus | 16 | 77 | 4.81 | 2.30 |
|  | 4 | H05..00 | Other acute upper respiratory infections | 50 | 217 | 4.34 | 6.49 |

|  | 5 | 173..00 | Breathlessness | 127 | 386 | 3.04 | 11.54 |
|  | 6 | M22..00 | Other dermatoses | 43 | 181 | 4.21 | 5.41 |
|  | 7 | 171..00 | Cough | 161 | 472 | 2.93 | 14.11 |
|  | 8 | 19F..00 | Diarrhoea symptoms | 64 | 225 | 3.52 | 6.72 |
|  | 9 | 1M1..00 | Pain in lower limb | 49 | 172 | 3.51 | 5.14 |
| Level 1-3 | 10 | F4C..00 | Disorders of conjunctiva | 43 | 184 | 4.28 | 5.50 |
|  | 11 | H06..00 | Acute bronchitis and bronchiolitis | 201 | 530 | 2.64 | 15.84 |
|  | 12 | M03..00 | Other cellulitis and abscess | 46 | 181 | 3.93 | 5.41 |
|  | 13 | AB0..00 | Dermatophytosis including tinea or ringworm | 22 | 102 | 4.64 | 3.05 |
|  | 14 | N21..00 | Peripheral enthesopathies and allied syndromes | 52 | 154 | 2.96 | 4.60 |
|  | 15 | N24..00 | Other soft tissue disorders | 229 | 573 | 2.50 | 17.12 |
|  | 16 | 168..00 | Tiredness symptom | 67 | 217 | 3.24 | 6.49 |
|  | 17 | 19C..00 | Constipation | 54 | 207 | 3.83 | 6.19 |
|  | 18 | 1A2..00 | Micturition control | 15 | 75 | 5.00 | 2.24 |
|  | 19 | 1A1..00 | Micturition frequency | 35 | 119 | 3.40 | 3.56 |
|  | 20 | F56..00 | Vertiginous syndromes, other disorders of vestibular system | 19 | 79 | 4.16 | 2.36 |

Table 6. The top 20 ADRs for Alendronate based on Wilcoxon rank-sum test.

|  | Rank | Readcodes | Medical events | NB | NA | R1(%) | R2 (%) |
|---|---|---|---|---|---|---|---|
|  | 1 | 1Z12.00 | Chronic kidney disease stage 3 | 51 | 189 | 3.71 | 5.65 |
|  | 2 | M0...00 | Skin and subcutaneous tissue infections | 9 | 65 | 7.22 | 1.94 |
|  | 3 | M03z000 | Cellulitis NOS | 37 | 152 | 4.11 | 4.54 |
|  | 4 | F4C0.00 | Acute conjunctivitis | 30 | 125 | 4.17 | 3.74 |
|  | 5 | G20..00 | Essential hypertension | 30 | 116 | 3.87 | 3.47 |
| Level 1-5 | 6 | 1M10.00 | Knee pain | 43 | 154 | 3.58 | 4.60 |
|  | 7 | C10F.00 | Type 2 diabetes mellitus | 7 | 44 | 6.29 | 1.32 |
|  | 8 | K190.00 | Urinary tract infection, site not specified | 62 | 214 | 3.45 | 6.40 |
|  | 9 | F46..00 | Cataract | 14 | 67 | 4.79 | 2.00 |
|  | 10 | H01..00 | Acute sinusitis | 21 | 78 | 3.71 | 2.33 |
|  | 11 | A53..11 | Shingles | 15 | 56 | 3.73 | 1.67 |
|  | 12 | J16y400 | Dyspepsia | 29 | 104 | 3.59 | 3.11 |
|  | 13 | 182..00 | Chest pain | 73 | 203 | 2.78 | 6.07 |
|  | 14 | K15..00 | Cystitis | 45 | 131 | 2.91 | 3.92 |
|  | 15 | N143.00 | Sciatica | 33 | 103 | 3.12 | 3.08 |
|  | 16 | 1D14.00 | C/O: a rash | 49 | 184 | 3.76 | 5.50 |
|  | 17 | H06z000 | Chest infection NOS | 86 | 297 | 3.45 | 8.88 |
|  | 18 | N05..11 | Osteoarthritis | 39 | 138 | 3.54 | 4.12 |
|  | 19 | N131.00 | Cervicalgia - pain in neck | 48 | 132 | 2.75 | 3.95 |
|  | 20 | 173I.00 | MRC Breathlessness Scale: grade 2 | 9 | 43 | 4.78 | 1.29 |
|  | 1 | 1Z1..00 | Chronic renal impairment | 58 | 237 | 4.09 | 7.08 |
|  | 2 | H05..00 | Other acute upper respiratory infections | 50 | 217 | 4.34 | 6.49 |
|  | 3 | M22..00 | Other dermatoses | 43 | 181 | 4.21 | 5.41 |
| Level 1-3 | 4 | C10..00 | Diabetes mellitus | 16 | 77 | 4.81 | 2.30 |
|  | 5 | H06..00 | Acute bronchitis and bronchiolitis | 201 | 530 | 2.64 | 15.84 |
|  | 6 | 183..00 | Oedema | 77 | 252 | 3.27 | 7.53 |
|  | 7 | M03..00 | Other cellulitis and abscess | 46 | 181 | 3.93 | 5.41 |
|  | 8 | 19F..00 | Diarrhoea symptoms | 64 | 225 | 3.52 | 6.72 |
|  | 9 | 173..00 | Breathlessness | 127 | 386 | 3.04 | 11.54 |
|  | 10 | K19..00 | Other urethral and urinary tract disorders | 82 | 297 | 3.62 | 8.88 |
|  | 11 | M0...00 | Skin and subcutaneous tissue infections | 9 | 65 | 7.22 | 1.94 |
|  | 12 | 171..00 | Cough | 161 | 472 | 2.93 | 14.11 |
|  | 13 | G20..00 | Essential hypertension | 35 | 130 | 3.71 | 3.89 |
|  | 14 | AB0..00 | Dermatophytosis including tinea or ringworm | 22 | 102 | 4.64 | 3.05 |
|  | 15 | F4C..00 | Disorders of conjunctiva | 43 | 184 | 4.28 | 5.50 |
|  | 16 | 1M1..00 | Pain in lower limb | 49 | 172 | 3.51 | 5.14 |
|  | 17 | 1A2..00 | Micturition control | 15 | 75 | 5.00 | 2.24 |
|  | 18 | 1A1..00 | Micturition frequency | 35 | 119 | 3.40 | 3.56 |
|  | 19 | F4D..00 | Inflammation of eyelids | 26 | 86 | 3.31 | 2.57 |
|  | 20 | 19C..00 | Constipation | 54 | 207 | 3.83 | 6.19 |

Table 7. The top 20 ADRs for Alendronate based on the descending order of R1 value.

|  | Rank | Readcodes | Medical events | NB | NA | R1 | R2 |
|---|---|---|---|---|---|---|---|
|  | 1 | F4K2100 | Vitreous detachment | 1 | 23 | 23.00 | 0.69 |
|  | 2 | N245.14 | Hand pain | 1 | 20 | 20.00 | 0.60 |
|  | 3 | K514.00 | Uterovaginal prolapse, unspecified | 0 | 18 | 18.00 | 0.54 |
|  | 4 | 16ZZ.00 | General symptom NOS | 1 | 18 | 18.00 | 0.54 |
|  | 5 | H040.00 | Acute laryngitis | 1 | 17 | 17.00 | 0.51 |
| Level 1-5 | 6 | N247012 | Swollen lower leg | 0 | 17 | 17.00 | 0.51 |
|  | 7 | 1955.00 | Heartburn | 1 | 17 | 17.00 | 0.51 |
|  | 8 | F4C0311 | Sticky eye | 1 | 17 | 17.00 | 0.51 |
|  | 9 | M22z.12 | Seborrhoeic wart | 1 | 17 | 17.00 | 0.51 |

|  | 10 | S8z..13 | Laceration | 1 | 16 | 16.00 | 0.48 |
|  | 11 | J046.00 | Temporomandibular joint disorders | 0 | 14 | 14.00 | 0.42 |
|  | 12 | 173H.00 | MRC Breathlessness Scale: grade 1 | 0 | 14 | 14.00 | 0.42 |
|  | 13 | SE4..11 | Leg bruise | 1 | 14 | 14.00 | 0.42 |
|  | 14 | 1B46.00 | C/O paraesthesia | 1 | 14 | 14.00 | 0.42 |
|  | 15 | G65..00 | Transient cerebral ischaemia | 1 | 14 | 14.00 | 0.42 |
|  | 16 | 1B16.11 | Agitated - symptom | 1 | 14 | 14.00 | 0.42 |
|  | 17 | 1B6..13 | Vasovagal symptom | 1 | 13 | 13.00 | 0.39 |
|  | 18 | 19A..00 | Abdominal distension symptom | 1 | 13 | 13.00 | 0.39 |
|  | 19 | H32..00 | Emphysema | 1 | 13 | 13.00 | 0.39 |
|  | 20 | 1AE..00 | Vaginal discomfort | 1 | 12 | 12.00 | 0.36 |
| Level 1-3 | 1 | J04..00 | Dentofacial anomalies | 0 | 16 | 16.00 | 0.48 |
|  | 2 | BB5..00 | [M]Adenomas and adenocarcinomas | 1 | 16 | 16.00 | 0.48 |
|  | 3 | A07..00 | Intestinal infection due to other organisms | 0 | 15 | 15.00 | 0.45 |
|  | 4 | N35..00 | Acquired deformities of toe | 0 | 14 | 14.00 | 0.42 |
|  | 5 | 19A..00 | Abdominal distension symptom | 1 | 14 | 14.00 | 0.42 |
|  | 6 | 1M0..00 | Pain in upper limb | 1 | 13 | 13.00 | 0.39 |
|  | 7 | H32..00 | Emphysema | 1 | 13 | 13.00 | 0.39 |
|  | 8 | 1J0..00 | Suspected malignancy | 0 | 12 | 12.00 | 0.36 |
|  | 9 | F11..00 | Other cerebral degenerations | 0 | 12 | 12.00 | 0.36 |
|  | 10 | B22..00 | Malignant neoplasm of trachea, bronchus and lung | 1 | 12 | 12.00 | 0.36 |
|  | 11 | 1AE..00 | Vaginal discomfort | 1 | 12 | 12.00 | 0.36 |
|  | 12 | 1CB..00 | Throat symptom NOS | 2 | 22 | 11.00 | 0.66 |
|  | 13 | A79..00 | Specific viral infections | 2 | 21 | 10.50 | 0.63 |
|  | 14 | M0z..00 | Skin and subcut tissue infection NOS | 2 | 21 | 10.50 | 0.63 |
|  | 15 | J02..00 | Pulp and periapical tissue disease | 1 | 10 | 10.00 | 0.30 |
|  | 16 | B46..00 | Malignant neoplasm of prostate | 1 | 10 | 10.00 | 0.30 |
|  | 17 | N36..00 | Other acquired limb deformity | 1 | 10 | 10.00 | 0.30 |
|  | 18 | SE2..00 | Contusion, trunk | 1 | 9 | 9.00 | 0.27 |
|  | 19 | BB3..00 | [M]Basal cell neoplasms | 1 | 9 | 9.00 | 0.27 |
|  | 20 | 15D..00 | Dyspareunia | 0 | 9 | 9.00 | 0.27 |

This table shows the top 20 results in descending order of R1 value for the detected results of Student's t-test with p<0.05.

Table 8. The top 20 ADRs for Metoclopramide based on Student's t-test.

|  | Rank | Readcodes | Medical events | NB | NA | R1(%) | R2 (%) |
|---|---|---|---|---|---|---|---|
| Level 1-5 | 1 | 22J..14 | Patient died | 2 | 101 | 50.50 | 1.21 |
|  | 2 | 22J..12 | Death | 1 | 72 | 72.00 | 0.87 |
|  | 3 | B590.00 | Disseminated malignancy NOS | 7 | 33 | 4.71 | 0.40 |
|  | 4 | H25..00 | Bronchopneumonia due to unspecified organism | 5 | 22 | 4.40 | 0.26 |
|  | 5 | 22J..13 | Died | 0 | 41 | 41.00 | 0.49 |
|  | 6 | 1737.00 | Wheezing | 2 | 13 | 6.50 | 0.16 |
|  | 7 | M0z..11 | Infected sebaceous cyst | 2 | 11 | 5.50 | 0.13 |
|  | 8 | 1D11.00 | C/O: a swelling | 1 | 10 | 10.00 | 0.12 |
|  | 9 | E2B..00 | Depressive disorder NEC | 34 | 73 | 2.15 | 0.88 |
|  | 10 | G65..00 | Transient cerebral ischaemia | 4 | 15 | 3.75 | 0.18 |
|  | 11 | 1B16.00 | Agitated | 3 | 14 | 4.67 | 0.17 |
|  | 12 | G573200 | Paroxysmal atrial fibrillation | 0 | 6 | 6.00 | 0.07 |
|  | 13 | 1A24.11 | Stress incontinence - symptom | 0 | 6 | 6.00 | 0.07 |
|  | 14 | J661.00 | Cholangitis | 0 | 6 | 6.00 | 0.07 |
|  | 15 | C320.00 | Pure hypercholesterolaemia | 4 | 16 | 4.00 | 0.19 |
|  | 16 | 2841.00 | Confused | 5 | 18 | 3.60 | 0.22 |
|  | 17 | F13z200 | Restless legs syndrome | 3 | 12 | 4.00 | 0.14 |
|  | 18 | 173I.00 | MRC Breathlessness Scale: grade 2 | 1 | 8 | 8.00 | 0.10 |
|  | 19 | M101.00 | Seborrhoeic dermatitis | 2 | 10 | 5.00 | 0.12 |
|  | 20 | 1732.00 | Breathless - moderate exertion | 0 | 7 | 7.00 | 0.08 |
| Level 1-3 | 1 | 22J..00 | O/E - dead | 3 | 221 | 73.67 | 2.66 |
|  | 2 | 1C1..00 | Hearing symptoms | 17 | 41 | 2.41 | 0.49 |
|  | 3 | F13..00 | Other extrapyramidal disease and abnormal movement disorders | 6 | 22 | 3.67 | 0.26 |
|  | 4 | H25..00 | Bronchopneumonia due to unspecified organism | 5 | 22 | 4.40 | 0.26 |
|  | 5 | K20..00 | Benign prostatic hypertrophy | 2 | 14 | 7.00 | 0.17 |
|  | 6 | A38..00 | Septicaemia | 3 | 16 | 5.33 | 0.19 |
|  | 7 | 284..00 | O/E - disorientated | 12 | 29 | 2.42 | 0.35 |
|  | 8 | SK1..00 | Other specified injury | 20 | 45 | 2.25 | 0.54 |
|  | 9 | E2B..00 | Depressive disorder NEC | 34 | 74 | 2.18 | 0.89 |
|  | 10 | C32..00 | Disorders of lipoid metabolism | 13 | 34 | 2.62 | 0.41 |
|  | 11 | H54..00 | Pulmonary congestion and hypostasis | 0 | 6 | 6.00 | 0.07 |
|  | 12 | F48..00 | Visual disturbances | 8 | 22 | 2.75 | 0.26 |
|  | 13 | K16..00 | Other disorders of bladder | 3 | 12 | 4.00 | 0.14 |
|  | 14 | M21..00 | Other atrophic and hypertrophic conditions of skin | 7 | 19 | 2.71 | 0.23 |
|  | 15 | N23..00 | Muscle, ligament and fascia disorders | 10 | 23 | 2.30 | 0.28 |
|  | 16 | B59..00 | Malignant neoplasm of unspecified site | 24 | 72 | 3.00 | 0.87 |

|  | 17 | F46..00 | Cataract | 11 | 27 | 2.45 | 0.32 |
|---|---|---|---|---|---|---|---|
|  | 18 | F4K..00 | Other eye disorders | 4 | 14 | 3.50 | 0.17 |
|  | 19 | 1A2..00 | Micturition control | 14 | 31 | 2.21 | 0.37 |
|  | 20 | M2y..00 | Other specified diseases of skin or subcutaneous tissue | 10 | 23 | 2.30 | 0.28 |

Table 9. The top 20 ADRs for Metoclopramide based on Wilcoxon rank-sum test.

|  | Rank | Readcodes | Medical events | NB | NA | R1(%) | 2 (%) |
|---|---|---|---|---|---|---|---|
| Level 1-5 | 1 | 22J..14 | Patient died | 2 | 101 | 50.50 | 1.21 |
|  | 2 | 22J..12 | Death | 1 | 72 | 72.00 | 0.87 |
|  | 3 | 22J..13 | Died | 0 | 41 | 41.00 | 0.49 |
|  | 4 | B590.00 | Disseminated malignancy NOS | 7 | 33 | 4.71 | 0.40 |
|  | 5 | D00..00 | Iron deficiency anaemias | 18 | 38 | 2.11 | 0.46 |
|  | 6 | E2B..00 | Depressive disorder NEC | 34 | 73 | 2.15 | 0.88 |
|  | 7 | H25..00 | Bronchopneumonia due to unspecified organism | 5 | 22 | 4.40 | 0.26 |
|  | 8 | E200300 | Anxiety with depression | 20 | 44 | 2.20 | 0.53 |
|  | 9 | F46..00 | Cataract | 9 | 23 | 2.56 | 0.28 |
|  | 10 | G580.11 | Congestive cardiac failure | 7 | 19 | 2.71 | 0.23 |
|  | 11 | C320.00 | Pure hypercholesterolaemia | 4 | 16 | 4.00 | 0.19 |
|  | 12 | 1B13.00 | Anxiousness | 13 | 30 | 2.31 | 0.36 |
|  | 13 | G65..00 | Transient cerebral ischaemia | 4 | 15 | 3.75 | 0.18 |
|  | 14 | B10..00 | Malignant neoplasm of oesophagus | 10 | 22 | 2.20 | 0.26 |
|  | 15 | M244.00 | Folliculitis | 7 | 18 | 2.57 | 0.22 |
|  | 16 | 2841.00 | Confused | 5 | 18 | 3.60 | 0.22 |
|  | 17 | 1B16.00 | Agitated | 3 | 14 | 4.67 | 0.17 |
|  | 18 | 1737.00 | Wheezing | 2 | 13 | 6.50 | 0.16 |
|  | 19 | M0z..11 | Infected sebaceous cyst | 2 | 11 | 5.50 | 0.13 |
|  | 20 | F13z200 | Restless legs syndrome | 3 | 12 | 4.00 | 0.14 |
| Level 1-3 | 1 | 22J..00 | O/E - dead | 3 | 221 | 73.67 | 2.66 |
|  | 2 | 1C1..00 | Hearing symptoms | 17 | 41 | 2.41 | 0.49 |
|  | 3 | C32..00 | Disorders of lipoid metabolism | 13 | 34 | 2.62 | 0.41 |
|  | 4 | F13..00 | Other extrapyramidal disease and abnormal movement disorders | 6 | 22 | 3.67 | 0.26 |
|  | 5 | B59..00 | Malignant neoplasm of unspecified site | 24 | 72 | 3.00 | 0.87 |
|  | 6 | F46..00 | Cataract | 11 | 27 | 2.45 | 0.32 |
|  | 7 | 1A2..00 | Micturition control | 14 | 31 | 2.21 | 0.37 |
|  | 8 | N23..00 | Muscle, ligament and fascia disorders | 10 | 23 | 2.30 | 0.28 |
|  | 9 | 284..00 | O/E - disorientated | 12 | 29 | 2.42 | 0.35 |
|  | 10 | E2B..00 | Depressive disorder NEC | 34 | 74 | 2.18 | 0.89 |
|  | 11 | H25..00 | Bronchopneumonia due to unspecified organism | 5 | 22 | 4.40 | 0.26 |
|  | 12 | SK1..00 | Other specified injury | 20 | 45 | 2.25 | 0.54 |
|  | 13 | F48..00 | Visual disturbances | 8 | 22 | 2.75 | 0.26 |
|  | 14 | M2y..00 | Other specified diseases of skin or subcutaneous tissue | 10 | 23 | 2.30 | 0.28 |
|  | 15 | M21..00 | Other atrophic and hypertrophic conditions of skin | 7 | 19 | 2.71 | 0.23 |
|  | 16 | K20..00 | Benign prostatic hypertrophy | 2 | 14 | 7.00 | 0.17 |
|  | 17 | A38..00 | Septicaemia | 3 | 16 | 5.33 | 0.19 |
|  | 18 | K16..00 | Other disorders of bladder | 3 | 12 | 4.00 | 0.14 |
|  | 19 | M10..00 | Erythematosquamous dermatosis | 9 | 20 | 2.22 | 0.24 |
|  | 20 | Eu4..00 | [X]Neurotic, stress - related and somoform disorders | 8 | 18 | 2.25 | 0.22 |

Table 10. The top 20 ADRs for Metoclopramide based on the descending order of R1 value.

|  | Rank | Readcodes | Medical events | NB | NA | R1(%) | R2 (%) |
|---|---|---|---|---|---|---|---|
|  | 1 | 22J..12 | Death | 1 | 72 | 72.00 | 0.87 |
|  | 2 | 22J..14 | Patient died | 2 | 101 | 50.50 | 1.21 |
|  | 3 | 22J..13 | Died | 0 | 41 | 41.00 | 0.49 |
|  | 4 | 1D11.00 | C/O: a swelling | 1 | 10 | 10.00 | 0.12 |
|  | 5 | 22J..00 | O/E - dead | 0 | 10 | 10.00 | 0.12 |
|  | 6 | A38..00 | Septicaemia | 1 | 9 | 9.00 | 0.11 |
|  | 7 | 173I.00 | MRC Breathlessness Scale: grade 2 | 1 | 8 | 8.00 | 0.10 |
|  | 8 | L05..12 | Termination of pregnancy | 1 | 8 | 8.00 | 0.10 |
|  | 9 | 19CZ.00 | Constipation NOS | 1 | 8 | 8.00 | 0.10 |
|  | 10 | 1732.00 | Breathless - moderate exertion | 0 | 7 | 7.00 | 0.08 |
|  | 11 | 2225.00 | O/E - dehydrated | 1 | 7 | 7.00 | 0.08 |
|  | 12 | A54..00 | Herpes simplex | 1 | 7 | 7.00 | 0.08 |
|  | 13 | S8z..13 | Laceration | 1 | 7 | 7.00 | 0.08 |
|  | 14 | M18z.11 | Skin irritation | 1 | 7 | 7.00 | 0.08 |
|  | 15 | A38z.11 | Sepsis | 1 | 7 | 7.00 | 0.08 |
|  | 16 | 22K5.00 | Body mass index 30+ - obesity | 1 | 7 | 7.00 | 0.08 |
|  | 17 | 1737.00 | Wheezing | 2 | 13 | 6.50 | 0.16 |
|  | 18 | G573200 | Paroxysmal atrial fibrillation | 0 | 6 | 6.00 | 0.07 |
|  | 19 | 1A24.11 | Stress incontinence - symptom | 0 | 6 | 6.00 | 0.07 |
|  | 20 | J661.00 | Cholangitis | 0 | 6 | 6.00 | 0.07 |
| Level 1-3 | 1 | 22J..00 | O/E - dead | 3 | 221 | 73.67 | 2.66 |
| | 2 | K20..00 | Benign prostatic hypertrophy | 2 | 14 | 7.00 | 0.17 |
| | 3 | H54..00 | Pulmonary congestion and hypostasis | 0 | 6 | 6.00 | 0.07 |
| | 4 | E13..00 | Other nonorganic psychoses | 1 | 6 | 6.00 | 0.07 |
| | 5 | K12..00 | Calculus of kidney and ureter | 1 | 6 | 6.00 | 0.07 |
| | 6 | A38..00 | Septicaemia | 3 | 16 | 5.33 | 0.19 |
| | 7 | H21..00 | Lobar (pneumococcal) pneumonia | 1 | 5 | 5.00 | 0.06 |
| | 8 | K56..00 | Noninflammatory vaginal disorders | 1 | 5 | 5.00 | 0.06 |
| | 9 | SK03.00 | Post-traumatic wound infection NEC | 1 | 5 | 5.00 | 0.06 |
| | 10 | A3Ay200 | Clostridium difficile infection | 0 | 5 | 5.00 | 0.06 |
| | 11 | 1CE..00 | C/O - wax in ear | 1 | 5 | 5.00 | 0.06 |
| | 12 | H25..00 | Bronchopneumonia due to unspecified organism | 5 | 22 | 4.40 | 0.26 |
| | 13 | K16..00 | Other disorders of bladder | 3 | 12 | 4.00 | 0.14 |
| | 14 | AC7..00 | Other intestinal helminthiases | 0 | 4 | 4.00 | 0.05 |
| | 15 | B6...00 | Malignant neoplasm of lymphatic and haemopoietic tissue | 0 | 4 | 4.00 | 0.05 |
| | 16 | L46..00 | Obstetric breast and lactation disorders NOS | 0 | 4 | 4.00 | 0.05 |
| | 17 | H52..00 | Pneumothorax | 2 | 8 | 4.00 | 0.10 |
| | 18 | S8z..00 | Open wound of head, neck and trunk NOS | 2 | 8 | 4.00 | 0.10 |
| | 19 | L05..00 | Legally induced abortion | 2 | 8 | 4.00 | 0.10 |
| | 20 | 17...00 | Respiratory symptoms | 1 | 4 | 4.00 | 0.05 |